\pdfoutput=1

\documentclass[11pt]{article}

\usepackage[]{acl}

\usepackage{times}
\usepackage{latexsym}

\usepackage[T1]{fontenc}

\usepackage[utf8]{inputenc}

\usepackage{microtype}

\usepackage{boldline,multirow}
\usepackage{colortbl}
\usepackage{makecell}
\usepackage{boldline,multirow}
\usepackage{colortbl}
\usepackage{makecell}
\usepackage[colorinlistoftodos]{todonotes}
\usepackage{soul} 
\usepackage{comment}
\usepackage{rotating, graphicx}

\usepackage{amsmath}
\usepackage{amssymb}

\definecolor{mygreen}{rgb}{0.0, 0.44, 0.0}

\newcommand{\astask}{AS2}
\newcommand{\tanda}{T{\sc and}A}

\newcommand{\dprast}{WQA}
\newcommand{\isa}{PEASI}
\newcommand{\isafull}{Passage-based Extracting Answer Sentence In-place}
\newcommand{\dprastfull}{Web-sourced Question Answering Dataset}
\newcommand{\easi}{EASI}
\newcommand{\easifull}{Extraction of Answer Sentence In-place}

\definecolor{mygreen}{rgb}{0.0, 0.44, 0.0}

\newcommand{\TANDA}{T{\sc AND}A}

\title{In Situ Answer Sentence Selection at Web-scale}
\author{Zeyu Zhang\thanks{\hspace{.5em}Work done while the author was an intern at Amazon Alexa AI.}\hspace{.3em},  Thuy Vu, \and  Alessandro Moschitti\\
  School of Information, The University of Arizona, Tucson, AZ, USA \\
  Amazon Alexa AI, Manhattan Beach, CA, USA \\
  \texttt{zeyuzhang@email.arizona.edu, \{thuyvu, amosch\}@amazon.com}}

\begin{document}
\maketitle
\begin{abstract}

Current answer sentence selection (AS2) applied in open-domain question answering (ODQA) selects answers by ranking a large set of possible candidates, i.e., sentences, extracted from the retrieved text.
In this paper, we present {\isafull} ({\isa}), a novel design for AS2 optimized for Web-scale setting, that, instead, computes such answer \emph{without} processing each candidate individually.
Specifically, we design a Transformer-based framework that jointly (i) reranks passages retrieved for a question and (ii) identifies a probable answer from the top passages in place.
We train {\isa} in a multi-task learning framework that encourages feature sharing between the components: passage reranker and passage-based answer sentence extractor.
To facilitate our development, we construct a new Web-sourced large-scale QA dataset consisting of 800,000+ labeled passages/sentences for 60,000+ questions.
The experiments show that our proposed design effectively outperforms the current state-of-the-art setting for AS2, i.e., a point-wise model for ranking sentences independently, by 6.51\% in accuracy, from 48.86\% to 55.37\%.
In addition, {\isa} is exceptionally efficient in computing answer sentences, requiring only $\sim$20\% inferences compared to the standard setting, i.e., reranking all possible candidates.
We believe the release of {\isa}, both the dataset and our proposed design, can contribute to advancing the research and development in deploying question answering services at Web scale.

\end{abstract}

\section{Introduction}


Open-domain Question Answering (ODQA) systems deliver answers to user questions directly in natural language, rather than returning a list of relevant documents as in traditional document search.
Such experience is becoming an industry standard in Web search, e.g., Google Search or Bing, and virtual assistants, e.g., Google Assistant or Alexa.
Two typical applications in ODQA are machine reading (MR) and answer sentence selection (AS2).
MR extracts an answer span from a relevant passage~\cite{kwiatkowski2019natural}, AS2 selects an answer sentence typically by ranking over all possible candidates~\cite{DBLP:conf/aaai/GargVM20}.
The latter is highly simulative of conversations in natural language, i.e., humans converse in compact and complete sentences.
We focus on AS2 in this paper by designing a novel paradigm for selecting answers at Web-scale efficiently.

\begin{figure}[t]
\center
\includegraphics[width=\linewidth]{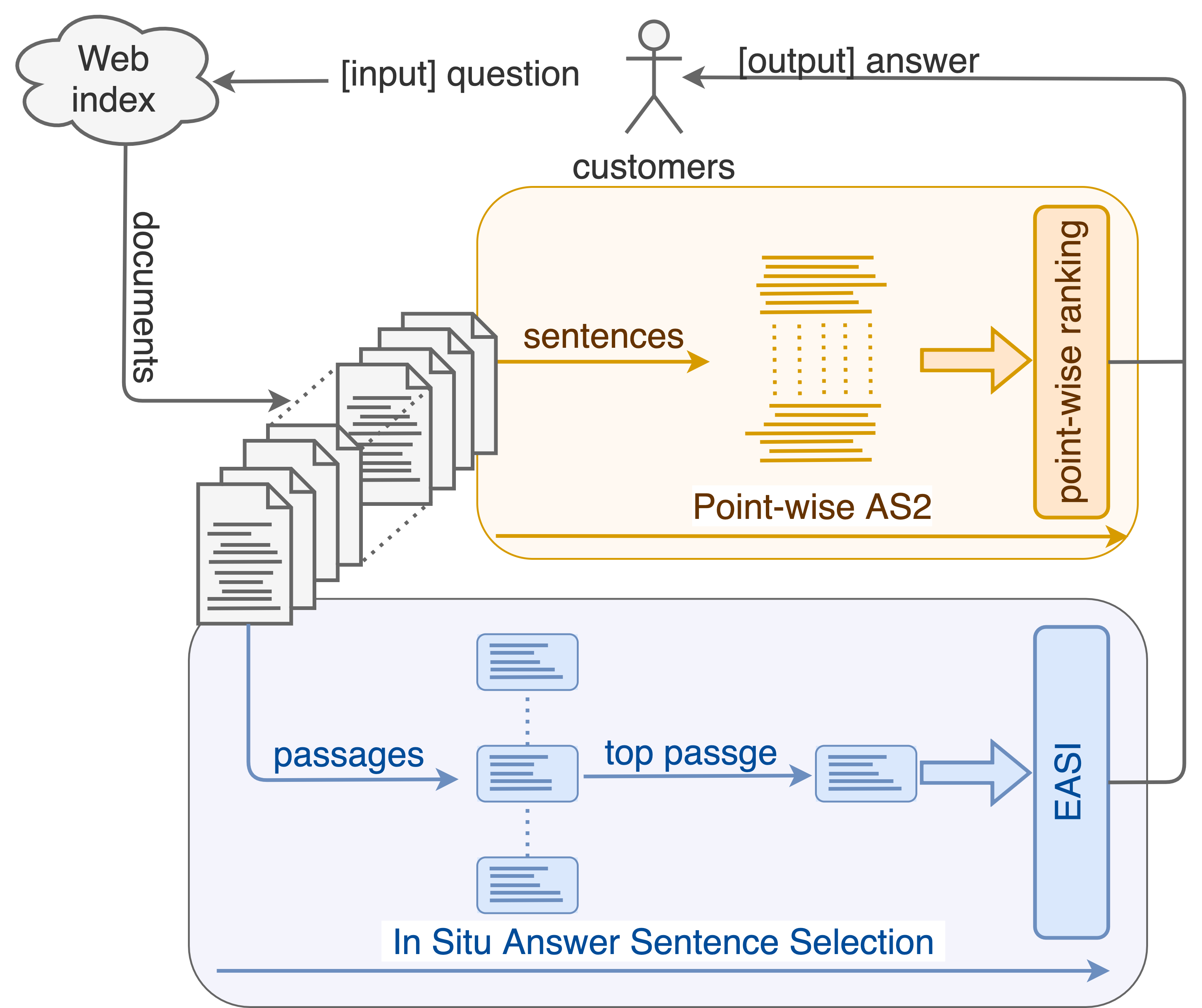}
\caption{Architecture of a standard point-wise answer sentence selection (the upper/orange pipeline) and our proposed {\isa}: an in situ answer sentence selection based on passage (the lower/blue pipeline).}
\label{as2-pipeline}
\end{figure}

In a nutshell, the current ODQA design for answer sentence selection (AS2) computes answer for a question in three phases: (i) text retrieval, which returns relevant content, documents or passages, for a question from a large text collection; (ii) text ranking, which reranks and decomposes text into answer candidates, e.g., sentences; and (iii) answer sentence selection (QA), which selects the final answer for a question from the list of candidates.
In practice, the pipeline may consist of several more components, e.g., question filtering~\cite{Garg2021WillTQ}, or employ sophisticated modeling arrangement~\cite{10.1145/3397271.3401266} to reduce cost for calculating the final answer.

From an engineering viewpoint, having many components would add I/Os overhead into the overall latency.
From a modeling perspective, it also hinders end-to-end optimization --- including feature sharing --- across components.
In particular, a passage reranker, for surfacing better relevant content to form a candidate set, and an answer sentence selector, for selecting the most probable answer sentence from the candidates, are separately built despite their modeling similarity.
In some cases, optimizing one component may even degrade the performance of the other.
Finally, performing point-wise ranking for candidates, i.e., the current state of the art~\cite{DBLP:conf/aaai/GargVM20}, is not effective as it completely ignores the passage semantic of an answer sentence in the modeling.
We address all these shortcomings in this paper.


Specifically, we design {\isafull} ({\isa}) for AS2 by consolidating the latter phases, i.e., all components after text retrieval, into a single framework.
{\isa} consists in two modules: a passage reranker (Pr) and an extractor of answer sentence in place for a passage (EASI); both are optimized jointly sharing features.
First, we propose to model the passage reranking step using Transformer-based models.
We employ a standard optimization technique for passage reranking while specializing the learning in surfacing potential passages for deriving correct answers.
Our passage reranker is robust in identifying potential passages for deriving an answer sentence.
Second, we introduce a novel setting for AS2 by selecting answer sentence \emph{directly} from a passage, instead of scoring each sentence from all relevant passages individually.
In particular, we design a multi-classifier architecture, leveraging the passage semantics, to identify the most probable answer sentence from a passage.
The setting allows to effectively identify better answer sentences compared standard setting.

In addition, {\isa} is also efficient as it can reduce significantly the computation required to derive an answer.
We train both components jointly in a multitask setting.
To facilitate the development of {\isa}, we develop a large-scale dataset of 64,000+ questions and 800,000+ labeled passages and sentences extracted from 30MM+ documents.
The experimental result shows that our proposed paradigm can effectively produce better answer sentences, improving 6.5\% in accuracy, from 48.9\% to 55.4\%, over the state-of-the-art model in AS2.
We also show the efficiency of our new setting, which reduces $\sim$81.4\% of computation required in a standard setting, i.e., reranking for all sentences.

In brief, the contribution of the paper is two-fold:
\begin{itemize}
  \item First, we construct a large dataset,~{\dprast}, for ODQA development which includes passage retrieval and answer reranking. The dataset establishes a new modeling task for ODQA using {\astask}.
  \item Second, we design a novel setting for {\astask}, namely {\isa}, that effectively delivers higher accurate answer via optimizing both the passage reranker and answer sentence identifier. We also show {\isa} to be efficient as it requires only $\sim$20\% inferences compared to point-wise AS2 setting.
\end{itemize}

We summarize the related work in Section~\ref{sec:related}.
We describe our proposed design and the construction of the dataset to facilitate the development in Section~\ref{sec:model} and Section~\ref{sec:data}, respectively.
We present the experimental studies in Section~\ref{sec:experiment}.
Section~\ref{sec:related} summarizes the related work.
Finally, we conclude our work in Section~\ref{sec:conclusion}.

\section{{\isa}: {\isafull}}
\label{sec:model}

In this section, we describe {\isa}, our proposed framework for deriving answer sentence for a large set of passages without processing each individual sentence.
{\isa} consists of two components, a passage reranker (PR) and a module extracting answer sentence in-place (EASI).

\subsection{Passage Reranker (PR)}
\label{sec:ranking}

A reranker can be simply built estimating the probability of correctness of a question and a passage, $p(q,p)$. 
Given a set of questions, $Q$, and a set of passages, $P$, we define a reranker as $R: Q \times \mathcal{P}(R) \rightarrow \mathcal{P}(R)$, which takes a subset $\Sigma$ of the sentences in $S$, and returns a subset of size $k<|\Sigma|$, i.e., $R(q,\Sigma) = (p_{i1},...,p_{ik})$, where $p(q,p_{ih})>p(q,p_{il})$, if $h<l$.
In brief, $q$ and $p$ are mapped to two sequences of column vectors corresponding to pre-trained word embeddings in $q$ and $p$, respectively.

We studied the efficient use of Transformer models, which are neural networks designed to capture dependencies between words, i.e., their interdependent context. 
The input is encoded in embeddings for tokens, segments, and their positions. These are given in input to several blocks (up to 24) containing layers for multi-head attention, normalization, and feedforward processing. The result of this transformation is an embedding, $\mathbf{x}$, representing a text pair that models the dependencies between words and segments of the two sentences.
We train these models for AS2 by feeding $\mathbf{x}$ to a fully connected layer,
and fine-tuning the Transformer as shown by \citet{devlin-etal-2019-bert}.
For this step, we use question-sentence pairs with labels being positive or negative if the sentence correctly answers the question or not.

\subsection{Extraction of Answer Sentence In-place (EASI)}

The current point-wise AS2 setting computes the best candidate via scoring for all possible candidates then ranking for the highest one.
The setting is efficient in scaling as the scoring function can be parallelized with more machines for latency.
However, the amount of scoring can be highly expensive, as it usually requires scoring thousands of candidates for a reasonable answer quality.
In our system, we propose a new setting for selecting answer sentences, in which we leverage the size constraint of passages indexed in dense-passage retrieval to optimize the extracting of answers.
In particular, we use a Transformer-based multi-classifier to select the most probable answer sentence given a passage.
We first concatenate a question with all of answer sentences from one passage, i.e., $\left(q [SEP] s_1 [SEP] s_2 \dots [SEP] s_{k}\right)$, where $k$ is the number of sentences for one passage.
We then feed the concatenation as input to the Transformer model.
We use the final hidden vector $E$ corresponding to the first input token $[CLS]$ generated by the Transformer, and a classification layer with weights $W \in R^{{(k)} \times |E|}$ to train the model.
We use a standard multi-class cross-entropy loss: $y \times log(softmax(EW^T))$, where $y$ is a one-hot vector representing labels for the $k$ sentences, i.e., $\left | y\right | = k$.
During the inference time, we calculate the scores of candidate answer sentences as $\big(p(c_1),..,p(c_{k})\big)=softmax(EW^T)$.
We then take the $c_i$ with the highest probability as final answer sentence.

\subsection{Multi-task Learning for {\isa}}

We train both PR and EASI using a multi-task architecture.
In this setting, both components can utilize the feature from the other component in modeling.
The training process is done in two stages.

In the first stage, we train each component individually as described in previous subsections.
In the second stage, we share the embedding features from EASI to PR.
Specifically, we first concatenate the question with each passage $p_i$ with two special tokens [CLS] and [SEP] to construct an input of $[CLS]q[SEP]p_i$.
The result of this transformation is an embedding, $\mathbf{E}$, representing $(q, p_i)$, which models the dependencies between the question and the passage.
In addition, we incorporate the embeddings from EASI to PR by concatenating it with the embedding for $(q, p_i)$.
We feed the concatenated embeddings, $2\mathbf{E}$ to a fully connected layer and one softmax to model the probability of the question/passage pair classification, as: $p(q,p_i) = softmax(W \times tanh(E(q,p_i), E(q, (c_1,...,c_k))]) + B)$. 
We train the model using binary cross-entropy loss: $\mathcal{L}=-\sum_{l \in \{0,1\}} y_l \times log(\hat{y}_l)$ on pairs of texts, where $y_l$ is the correct and incorrect answer label, $\hat{y}_1 = p(q,p_i)$, and $\hat{y}_0 = 1-p(q,p_i)$.

\section{{\dprast} Dataset Development for {\isa}}
\label{sec:data}

We describe our strategy to develop a general dataset for AS2 while retaining the connection of labeled answer sentences with their passages.
We name the dataset {\dprastfull} ({\dprast}) as we only use Web's data as material to construct the dataset.

\subsection{Question-Passage Relevance Definition}
Data construction for supervised passage reranking has been optimized mainly for machine reading (MR) task~\cite{lee2019latent,joshi-etal-2017-triviaqa,DBLP:journals/corr/abs-2012-01414}.
In this setting, relevance between a question and a passage is defined based on the presence of an answer span, the target in modeling an MR component.
The consideration is sub-optimal due to the high possibility of (1) false negatives, i.e., there can be infinite variants of a correct answer span to be captured sufficiently in the golden set, and (2) false positives, i.e., a presence of a correct answer span does not guarantee valid~\cite{DBLP:journals/corr/abs-2101-00133}.
Most importantly, it also ignores non-factoid questions, the major question type in ODQA.

In point-wise AS2 setting, we define a passage to be relevant to a question if it contains a correct answer sentence for the question.
Our definition is highly conservative compared to previous datasets as it requires direct answerability relation between a question and a passage.
Formally, let $p$ be a passage consisting of $n$ sentences $p=\{s_1, s_1, \dots, s_n\}$.
We define a question-passage pair $\left(q,p\right)$ relevant if there exists $s_i \in p$ such that $\left(q, s_i\right)$ is a correct question-answer pair.

\subsection{Annotating Question-Answer Relevance}

We describe our annotation criteria and the proposed pipeline which generates data for our annotation.
In retrieval-based ODQA, an answer to a question can be extracted from any Web document.
To this end, we build a standard ODQA pipeline that enables us to select highly relevant candidates from a large set of sentences retrieved by a Web index.
The questions and their top candidates are validated by annotator experts, judging the quality of a question and an answer candidate to a question.

\paragraph{Question Sampling} 
We sample the questions from a large collection of information inquiries collected from Web data, e.g., CommonCrawl.
We only consider well-formed questions.

\paragraph{Question-Answer Annotation} For each answer candidate of a question, we annotate if it correctly and sufficiently answers the question.
As we aim for high-quality data, a correct answer to a question must address the inquired information and also be conversational and direct to the question.

\subsection{{\dprast} Dataset Construction}
\label{data:development}
We build a standard ODQA to generate data for constructing the~{\dprastfull} ({\dprast}).
We also describe our strategy to propagate labels from answer sentences back to passages and documents to develop the dataset.

\paragraph{Web Document Collection} First and foremost, we build a large set of Web documents to facilitate the development of the paper, including data construction and system evaluation.
This resource allows us to measure the impact of our work in an ODQA industry-scale setting.
We selected English Web documents of 5,000 most popular domains, including Wikipedia, from the recent releases of Common Crawl of 2019 and 2020.
We then filtered pages that are too short or without proper HTML structures, i.e., having title and content.
The pages are then split into passages using the procedure described in DPR~\cite{Karpukhin20dense}.
This process produced a collection of $\sim$100MM and $\sim$130MM of documents and passages, respectively.

\paragraph{Web Document Index} 
We build two Web-scale document indexes using the Web Document Collection described above: a standard index using Lucene/Elasticsearch and a neural-based index using DPR~\cite{Karpukhin20dense}.
We will make this document collection available to the community to encourage future benchmarking of retrieval models, including lexical-based and neural-based, in an industrial setting.

\paragraph{Annotation Data Generation}
For each question, we retrieved up to 1,000 documents using both indexes.
We split the documents into sentences and applied a public state-of-the-art AS2 model~\cite{DBLP:conf/aaai/GargVM20} to select top candidates for annotation.
For each question, we annotate the top 15 candidate sentences.

\paragraph{Sentence to Passage Label Propagation}
For each annotated answer sentence, we select all passages that contain the sentence.
We label a passage positive if it contains at least one correct answer sentence.
The practice creates passages with answerability labels for all answerable questions.
The length of a passage is limited to 200 tokens.
We also do a sliding window to create more passages of the size that contain the candidate sentence.
This helps generate more passages with labels for the dataset.

\begin{figure}[t]
\center
\includegraphics[width=\linewidth]{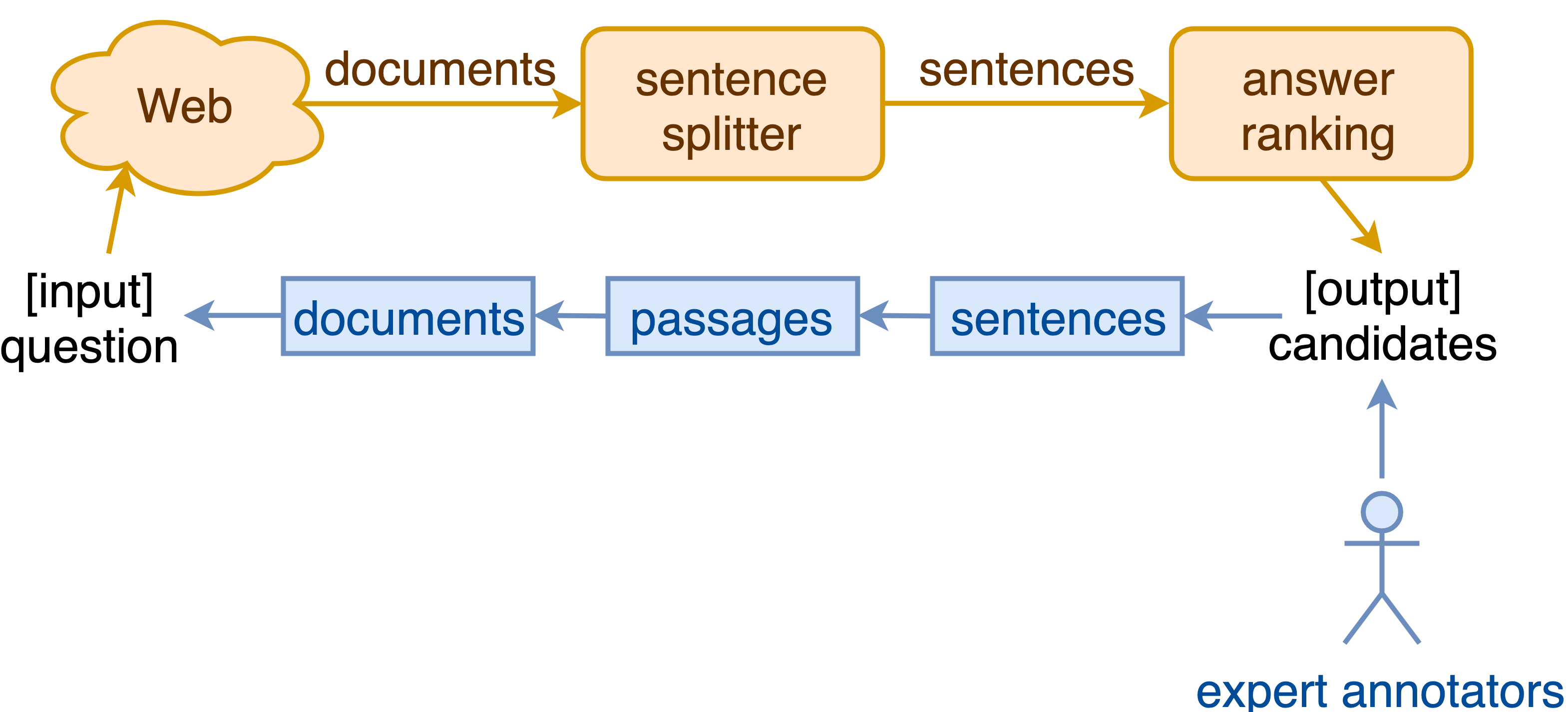}
\caption{This diagram describes the ODQA pipeline which generates question-answer pairs for annotation (the upper/orange pipeline), starting from an input question and ending with candidates. The (question, candidates) are sent to expert annotators for labeling. The labels of sentences are propagated back to the embedding passages and documents (the lower/blue pipeline).}
\label{pipeline}
\end{figure}

Figure~\ref{pipeline} describes the diagram of generating annotation data and constructing labeled data for ODQA.
Finally, Table~\ref{tbl:ourdataset} summarizes the statistics of our {\dprast} dataset.

\begin{table*}[]
\centering
\resizebox{1\linewidth}{!}{
\begin{tabular}{|r|r|r|rrr|rrr|}
\hline
\multicolumn{1}{|c|}{\multirow{2}{*}{\textbf{Split}}} & \multicolumn{1}{c|}{\multirow{2}{*}{\textbf{\#Questions}}} & \multicolumn{1}{c|}{\multirow{2}{*}{\textbf{\#Documents}}} & \multicolumn{3}{c|}{\textbf{\#Passages}}                                                                             & \multicolumn{3}{c|}{\textbf{\#Answers}}                                                                              \\ \cline{4-9} 
\multicolumn{1}{|c|}{}                                & \multicolumn{1}{c|}{}                                      & \multicolumn{1}{c|}{}                                      & \multicolumn{1}{c|}{\textbf{\#QP Pairs}} & \multicolumn{1}{c|}{\textbf{\#QP+}} & \multicolumn{1}{c|}{\textbf{\#QP-}} & \multicolumn{1}{c|}{\textbf{\#QA Pairs}} & \multicolumn{1}{c|}{\textbf{\#QA+}} & \multicolumn{1}{c|}{\textbf{\#QA-}} \\ \hline
\textbf{Train}                                        & 53,419                                                     & 18,983,495                                                 & \multicolumn{1}{r|}{695,033}             & \multicolumn{1}{r|}{265,177}        & 429,856                             & \multicolumn{1}{r|}{695,130}             & \multicolumn{1}{r|}{265,217}        & 429,913                             \\ \hline
\textbf{Dev}                                          & 5,416                                                      & 6,015,441                                                  & \multicolumn{1}{r|}{70,981}              & \multicolumn{1}{r|}{26,989}         & 43,992                              & \multicolumn{1}{r|}{70,988}              & \multicolumn{1}{r|}{26,992}         & 43,996                              \\ \hline
\textbf{Test}                                         & 5,395                                                      & 6,042,851                                                  & \multicolumn{1}{r|}{70,365}              & \multicolumn{1}{r|}{26,574}         & 43,791                              & \multicolumn{1}{r|}{70,372}              & \multicolumn{1}{r|}{26,576}         & 43,796                              \\ \hline
\hline
\textbf{Total}                                        & 64,230                                                     & 31,041,787                                                 & \multicolumn{1}{r|}{836,379}             & \multicolumn{1}{r|}{318,740}        & 517,639                             & \multicolumn{1}{r|}{836,490}             & \multicolumn{1}{r|}{318,785}        & 517,705                             \\ \hline
\end{tabular}
}
\caption{Statistics of our developed {\dprastfull} ({\dprast}) dataset.}
\label{tbl:ourdataset}
\end{table*}

\section{Experiments}
\label{sec:experiment}

In this section, we study the performance of the proposed {\isa}.
First, we describe the experiment settings for the section.
Second, we study the effectiveness of {\isa} by evaluating the performance for each separate component and for the whole design together.
We then analyze the efficiency of {\isa} by measuring the cost with a standard point-wise AS2 setting.
Finally, we analyze the result qualitatively.

\subsection{Experiment Setting}
\label{sec:setting}

\paragraph{Evaluation Metrics} We measure the performance a ranking result for a set of passages or sentences, and a final answer output from a pipeline using the following three metrics:
\begin{itemize}
  \item Precision at 1 (Accuracy): Percentage of questions having relevant passage or correct answer sentence returned at top 1. The measure computes the accuracy of a component.
  \item Mean Average Precision (MAP) and Mean Reciprocal Rank (MRR) for the mean of average precision scores and reciprocal ranks for a question set, respectively.
\end{itemize}

\paragraph{Dataset} We use our {\dprast} for our experiments as there is no equivalent public dataset available.
We will release {\isa}, both the dataset and our proposed design, to contribute to the research and development in deploying question answering services at Web scale.

\paragraph{Training Configuration}

For PR, we use 8 Nvidia-A100 GPU with batch size set at  64 per GPU and the Adam optimizer with a learning rate of $5e\text{-}6$ for training. The max sequence length is 512 and the number of epochs is 3.
In the multi-task training, we increase the number of epoch to 10.
For EASI, we also use the same training configuration as in PR, but use 20 epochs in the training.
Both components use RoBERTa-base Transformer models.
However, for PR, we use the public transferred model from~\cite{DBLP:conf/aaai/GargVM20}: TANDA-RoBERTa-base-ASNQ.

\subsection{Passage Reranking Evaluation}

In this section, we study the efficiency of the proposed. 
In particular, we compare the following training setting for the passage reranker.

\begin{itemize}
	\item {\bf Baseline}: PR is tuned with a standard Transformer
	\item {\bf MTL$_0$(PR, EASI)}: PR is tuned together with EASI and sharing the Transformer layers 
	\item {\bf MTL$_1$(PR, EASI)}: PR is tuned together with EASI with different Transformer
	\item {\bf MLT$_1$(PR $\bigoplus$ EASI)}: PR is tuned together with EASI and incorporate the concatenated embeddings from EASI
\end{itemize}

We also show the performance of TANDA in AS2 setting for a comparison in terms of accuracy, i.e., P@1.

\begin{table}[]
\resizebox{1\linewidth}{!} {
\begin{tabular}{|r|r|r|r|}
\hline
\multicolumn{1}{|c|}{\textbf{Model}} & \multicolumn{1}{c|}{\textbf{P@1}} & \multicolumn{1}{c|}{\textbf{MAP}} & \multicolumn{1}{c|}{\textbf{MRR}} \\ \hline
AS2/TANDA~\cite{DBLP:conf/aaai/GargVM20}                             & 0.489                             & 0.549                             & 0.556                             \\ \hline
\hline
Baseline Passage Reranker (PR)                             & 0.545                             & 0.575                             & 0.642                             \\ \hline
MTL$_0$(PR, EASI)          & 0.535                             & 0.569                             & 0.634                             \\ \hline
MTL$_1$(PR, EASI)          & 0.551                             & 0.575                             & 0.644                             \\ \hline
MLT$_1$(PR $\bigoplus$ EASI)        & 0.554                             & 0.576                             & 0.648                             \\ \hline
\end{tabular}
}
\caption{Multitask Performance}
\label{tbl:mtl}
\end{table}

The result shows that the baseline PR model outperforms AS2/TANDA setting by a large margin. 
This is expected as we can learn more discriminative features from passages compared to sentences.
In MTL setting,  {\bf MTL$_0$(PR, EASI)} has a drop in performance as we try to use the same Transformer-model for both PR and EASI.
This suggests that while both models encode a passage for a downstream task, they are not optimized identically.
Second, the improved result of {\bf MTL$_1$(PR, EASI)} indicates the effectiveness in learning both tasks together, adding 0.6\% in accuracy.
Finally, we achieve another 0.3\% improvement when directly sharing embedding features across tasks.

\subsection{{\easifull} ({\easi}) Evaluation}

In this section, we measure the performance of EASI in the context of answer sentence selection for ODQA.

First, we train EASI on the WQA dataset described in Section~\ref{sec:data} using only passages that have correct answers.
This is based on the reasoning that EASI is built specifically to identify a correct answer when there is one.
One important observation is that, unlike standard point-wise AS2 setting, EASI setting is \emph{extremely} simpler as it requires to identify the correct answer from a few candidates, i.e., sentences in a passage.
Therefore, we can easy reach the performance for EASI at 99.98\%.
We also use {\dprast} dataset to build the model for {\TANDA}, the current state-of-the-art AS2 model~\cite{DBLP:conf/aaai/GargVM20}.

We compare the performance in two designs: standard AS2 and {\isa}.
For AS2, we apply the model to score all candidate sentences found in the passages.
We consider two modeling settings of {\TANDA}:
\begin{itemize}
\item {\bf Public {\TANDA}}: this is the original model released in the paper~\cite{DBLP:conf/aaai/GargVM20}.
\item {\bf {\TANDA} fine-tuned on {\dprast}}: we continue fine-tuned the public model on our dataset. This is the evaluation baseline.
\end{itemize}

For EASI, we first use a PR model to rank the passages before feeding into the answer selection step.
We consider two settings:
\begin{itemize}
\item {\bf {\isa} for all passages $\rightarrow$ {AS2}}: we select the best candidate from each passage by running EASI for all passages. We then apply {\TANDA} to select the best answer among the top passages.
\item {\bf {\isa} for top-1 passage}: we apply EASI to the top-1 passage selected by the PR and computes the candidate directly.
\end{itemize}

As described in Section~\ref{sec:data}, each labeled answer would generate approximately 5 passages having the position of the labeled answer from 1 to 5.
Therefore, we also study the performance of~{\isa} in three groups of passages: (i) all generated passages ({\bf All}), (ii) all passages where the labeled sentence is in the middle of the passage ({\bf Center}), and (iii) selecting a random passage, only one, for each labeled sentence ({\bf Random}).
Table~\ref{tbl:ase-perf} shows the result on {\dprast}'s dev and test sets.

\begin{table*}
\centering
\resizebox{1\linewidth}{!} {
\begin{tabular}{|r|r|l|r|r|r|r|r|r|} 
\hline
\multicolumn{1}{|c|}{\multirow{2}{*}{\bf Design}} & \multicolumn{2}{c|}{\multirow{2}{*}{\bf Pipeline Setting}}                     & \multicolumn{3}{c|}{\bf Dev}                                                       & \multicolumn{3}{c|}{\bf Test}                                                       \\ 
\cline{4-9}
\multicolumn{1}{|c|}{}                        & \multicolumn{2}{c|}{}                                                      & \multicolumn{1}{c|}{\bf P@1} & \multicolumn{1}{c|}{\bf MAP} & \multicolumn{1}{c|}{\bf MRR} & \multicolumn{1}{c|}{\bf P@1} & \multicolumn{1}{c|}{\bf MAP} & \multicolumn{1}{c|}{\bf MRR}  \\ 
\hline
\multirow{2}{*}{AS2}    & \multicolumn{2}{l|}{[Public] {\TANDA}~\cite{DBLP:conf/aaai/GargVM20}}                            & \multicolumn{1}{l|}{0.399} & \multicolumn{1}{l|}{0.468} & 0.495 & \multicolumn{1}{l|}{0.396} & \multicolumn{1}{l|}{0.464} & 0.490 \\ \cline{2-9} 
                        & \multicolumn{2}{l|}{[Baseline] {\TANDA} fine-tuned on {\dprast}} & \multicolumn{1}{l|}{0.485} & \multicolumn{1}{l|}{0.548} & 0.551 & \multicolumn{1}{l|}{0.489} & \multicolumn{1}{l|}{0.549} & 0.556 \\ \hline
\multirow{6}{*}{{\isa}}                        & \multirow{2}{*}{All}            & {\isa} for all passages $\rightarrow$ {AS2} & 0.529                    & 0.586                    & 0.584                    & 0.537                    & 0.588                    & 0.592                     \\ 
\cline{3-9}
                                              &                                 & {\isa} for top-1 passage                     & 0.559                    & -                        & -                        & 0.545                    & -                        & -                         \\ 
\cline{2-9}
                                              & \multirow{2}{*}{Center} & {\isa} for all passages $\rightarrow$ {AS2} & 0.529                    & 0.603                    & 0.641                    & 0.537                    & 0.607                    & 0.648                     \\ 
\cline{3-9}
                                              &                                 & {\isa} for top-1 passage                     & 0.549                    & -                        & -                        & 0.548                    & -                        & -                         \\
\cline{2-9}
                                              & \multirow{2}{*}{Random}         & {\isa} for all passages $\rightarrow$ {AS2} & 0.529                    & 0.603                    & 0.641                    & 0.537                    & 0.607                    & 0.648                     \\ 
\cline{3-9}
                                              &                                 & {\isa} for top-1 passage                     & 0.556                    & -                        & -                        & 0.551                    & -                        & -                         \\ 
\hline
\end{tabular}
}
\caption{Performance of {\isa} compared with the current state-of-the-art AS2~\cite{DBLP:conf/aaai/GargVM20}.}
\label{tbl:ase-perf}
\end{table*}

For AS2 setting, the result shows that we have a good baseline by continue fine-tuning the public TANDA using our dataset.
We find the performance of all {\isa} settings to be better than traditional AS2 model tuned via TANDA, the state-of-the-art in point-wise answer reranking.
In addition, the result shows that applying {\isa} only once, on top-1 passage, outperforms a more complicated pipeline that processes for all passages and reranks again with TANDA.
While EASI returns the best candidate from each passage, point-wise model may degrade the performance by making predictions without having the semantic of the passages.
Finally, the result shows comparable performance across different groups of passages. 
This empirically confirms that {\isa} can effectively identify a good answer from a relevant passage.

\subsection{Cost Analyses}

We analyze the cost efficiency for operating standard AS2, e.g., using {\tanda}~\cite{DBLP:conf/aaai/GargVM20}, and our proposed~\isa.
We compare with {\tanda} since it is the current state-of-the-art point-wise model for AS2.
We show the cost for overall~{\isa}, {\isa:ALL}, and for each component, {\isa}:PR and {\isa}:EASI.
In each model setting, we present the prediction number, i.e, number of inferences, the average cost per prediction, and the overall latency (in millisecond) measured on the same environment.
Table~\ref{tbl:efficiency} shows the results.

\begin{table}[]
\resizebox{\linewidth}{!}{%
\begin{tabular}{|l|r|r|r|}
\hline
\multicolumn{1}{|c|}{\bf Model} & \multicolumn{1}{c|}{\bf \#Predictions} & \multicolumn{1}{c|}{\bf Cost (ms)} & \multicolumn{1}{c|}{\bf Latency (ms)} \\ \hline
AS2/{\tanda}                       & 215                              & 11.7                                      & 2,521                             \\ \hline
{\isa}:PR                          & 43                               & 10.9                                      & 469                               \\ \hline
{\isa}:EASI                          & 1                                & 10.0                                      & 10                                \\ \hline
{\isa:ALL}                     & 44                               & 10.9                                      & 479                               \\ \hline
\end{tabular}
}
\caption{Efficiency analysis for processing each question}
\label{tbl:efficiency}
\end{table}

The result shows that the cost for each prediction is relatively similar.
This is reasonable since they are both based on standard Transformer-based model.
Therefore, {\isa} is significantly efficient as it requires only a fifth of the number of predictions required in a standard AS2 setting.
This is because {\isa} processes a volume of several sentences at once rather than for each sentence.
As studied in previous section, the consideration also allows to leverage the passage semantic more effectively, enabling higher accuracy in delivering the best answer.

\subsection{Qualitative Analysis}

We analyze qualitatively two example questions in Table~\ref{tbl:example-2} and Table~\ref{tbl:example-3}, in which {\isa} outperforms {\TANDA} to understand the behavior of each setting.

\begin{table}[h]
\centering
\resizebox{\linewidth}{!}{%
\begin{tabular}{|lp{8cm}|}
\hline
$question$: & \textbf{What is the biggest fish in the world?}\\
& \\
$a_{\TANDA}$: &\vspace{-0.8em} {\bf \color{red} The biggest fish in the water is the massive taimen, the largest species of trout on the planet.}\\
$p_{a_{\TANDA}}$: &\vspace{-0.8em} {Mexico has a number of great fresh and saltwater destinations for fishing, but perhaps the best can be found in {\color{red}Loreto on the Baja Peninsula}. The region is home to an array of fantastic species of sport fish, including sea bass, yellowfin tuna, and snapper, all of which are large, energetic, and fun to reel in. {\color{red}The burgeoning fishing scene in Loreto} has attracted top guides who know the area well and can help visitors get the most out of their trip. Just dress for warm conditions, bring plenty of sunscreen, and be prepared for some great days out on the water. {\bf \color{red} The biggest fish in the water is the massive taimen, the largest species of trout on the planet.}}\\
$a_{\isa}$: &\vspace{-0.8em}{\bf \color{mygreen} The largest fish is the Whale Shark.}\\
$p_{a_{\isa}}$: &\vspace{-0.8em} The largest animal {\color{mygreen}on the planet} is the Blue Whale. {\bf \color{mygreen} The largest fish is the Whale Shark.}  Both of these marine mammals are filter feeders. {\color{mygreen}The ocean is home to some of the weirdest, creepiest and largest living creatures on this planet}. A mere glimpse of them spawned countless tales of sea monsters, such as the Loch Ness Monster and the Kraken.\\

\hline
\end{tabular}
}
\caption{An example shows {\isa} to select a better an answer for a question by leveraging the passage information.}
\label{tbl:example-2}
\end{table}

In the first example, Table~\ref{tbl:example-2}, we find that TANDA selects a wrong answer with a very plausible phrasing, i.e., ``The biggest fish in the water is ...''.
The answer is incorrect because ``the biggest fish'' mentioned in the selected answer is narrowed within the ``Loreto'' area. 
However, this semantic is not incorporated into the modeling.
On the other hand, {\isa} correctly selects a good answer even though the sentence itself is minimal.
In this case, by observing the embedding passage, we can easily see that the passage is indeed referring to the information asked in the question.
The analysis confirms that {\isa} can effectively capture the semantic of the passage and robustly identify the most probable answer to the question.

\begin{table}[t]
\centering
\resizebox{\linewidth}{!}{%
\begin{tabular}{|lp{8cm}|}
\hline
$question$: & \textbf{Why do volcanic eruptions occur in Hawaii?}\\
& \\
$a_{\TANDA}$: &\vspace{-0.8em}{\bf \color{red}Hawaiian eruptions are considered less dangerous than other types of volcanic eruptions, due to the lack of ash and the generally slow movement of lava flows.}\\
$p_{a_{\TANDA}}$: & {These rocks are similar but not identical to those that are produced at ocean ridges. Basalt relatively richer in sodium and potassium (more alkaline) has erupted at the undersea volcano of Lōihi at the extreme southeastern end of the volcanic chain, and these rocks may be typical of early stages in the "evolution" of all Hawaiian islands. In the late stages of eruption of individual volcanoes, more alkaline basalt also was erupted, and in the very late stages after a period of erosion, rocks of unusual composition such as nephelinite were produced in very small amounts. These variations in magma composition have been investigated in great detail, in part to try to understand how mantle plumes may work. {\bf \color{red}Hawaiian eruptions are considered less dangerous than other types of volcanic eruptions, due to the lack of ash and the generally slow movement of lava flows.}}\\
& \\
$a_{\isa}$: &\vspace{-0.8em}{\bf \color{mygreen}Hawaiian eruptions usually start by the formation of a crack in the ground from which a curtain of incandescent magma or several closely spaced magma fountains appear.}\\
$p_{a_{\isa}}$: &\vspace{-0.8em} In fissure-type eruptions, lava spurts from a fissure on the volcano's rift zone and feeds lava streams that flow downslope. In central-vent eruptions, a fountain of lava can spurt to a height of 300 meters or more (heights of 1600 meters were reported for the 1986 eruption of Mount Mihara on Izu Ōshima, Japan). {\bf \color{mygreen}Hawaiian eruptions usually start by the formation of a crack in the ground from which a curtain of incandescent magma or several closely spaced magma fountains appear.}  {\color{mygreen}The lava can overflow the fissure and form aā or pāhoehoe style of flows. When such an eruption from a central cone is protracted, it can form lightly sloped shield volcanoes, for example Mauna Loa or Skjaldbreiður in Iceland.}\\
& \\
\hline
\end{tabular}
}
\caption{An example shows {\isa} to select a better an answer for a question by leveraging the passage information.}
\label{tbl:example-3}
\end{table}

Table~\ref{tbl:example-3} shows another example for a ``why'' question, a difficult case in QA that requires reasoning.
We again find that TANDA gets wrong because it incorrectly relies on a phrasing pattern aiming to provide reasoning, i.e., ``due to ...'' in a selected sentence about ``Hawaiian eruptions.''
Both failures, despite the difference, emphasize the shortcoming of point-wise modeling, in which an answer is selected merely based on a plausible phrasing, rather than addressing the question properly.
While {\isa} correctly selects a good answer even though the sentence is not explicitly phrased for an explanation.
In this case, we observe that other sentences in the passage contribute to selecting a sound answer.

\section{Related Work}
\label{sec:related}

Our work aims at improving the effectiveness and efficiency in deploying AS2 at Web scale, with answers computed from documents or passages retrieved from a large Web index.

Regarding effectiveness, the task of reranking answer-sentence candidates provided by a retrieval engine is traditionally modeled with a classifier scoring the candidates.
Previous work targeting ranking problems in the text domain has classified reranking functions into three buckets: pointwise~\cite{shen-etal-2017-inter,DBLP:journals/corr/abs-1905-12897,DBLP:conf/aaai/GargVM20}, pairwise~\cite{conf/cikm/RaoHL16,tayyar-madabushi-etal-2018-integrating,laskar-etal-2020-contextualized}, and listwise methods~\cite{cao2007learning,conf/cikm/Bian0YCL17,DBLP:journals/corr/abs-1804-05936}.
Regarding efficiency, there have been recent works focusing on optimizing for AS2 deployment at Web scale.
A general approach is to reduce the candidate set via filtering.
For example, ~\cite{10.1145/3397271.3401266} proposed to improve the candidate set by extending the search space, using more passages, then filtering out less probable candidates by using a sequence of lower-cost models.
While the result shows good improvement, the system implementation would require significant engineering effort.
In addition, the interaction among several models would increase I/O overhead into the overall latency.

In our paper, we identify that the current setting, treating each answer sentence individually and ignoring the passage semantic, is not optimal.
We first perform reranking at passage level, instead of candidate level, and then select the best candidate directly from the top passage.
We showed in our paper that utilizing the passage semantic improves the performance of answer identification significantly.
In addition, as we only perform the reranking at passage level, we can increase the efficiency by reducing $\sim$80\% inferences.

\medskip

Besides, our work is also related to several applications of multitask learning (MTL) in NLP~\cite{sogaard-goldberg-2016-deep,rei-2017-semi}.
In question answering, MTL has also been studied to perform several related tasks.
In answer generation,~\citet{choi-etal-2017-coarse} proposed to learn a sentence selection and answer generation model jointly.
For machine reading task,~\citet{DBLP:journals/corr/abs-1709-00023} jointly train a ranking and reader model for ODQA.
To the best of our knowledge, we are the first to apply multitask learning for AS2.

\section{Conclusion}
\label{sec:conclusion}

We presented a novel design for selecting the answer sentence directly from a passage to facilitate Web-scale open-domain question answering (ODQA).
Specifically, we first describe our proposed~{\isafull} ({\isa}) that consists a passage reranker and answer sentence extractor for a passage.
The components are trained together in a multi-task setting.
In addition, we explain our strategy in developing a general QA dataset that also facilitates the development of {\isa}.
The dataset contains several pieces of information collected in a standard end-to-end ODQA setting, including questions, retrieved documents, passages, and answer sentence candidates with human labeling given the question context.
The dataset is of large scale, having $\sim$800K annotations for answer candidates for $\sim$65K questions.
The candidates are extracted from ~30MM passages, retrieved from a large index having 130MM passages from CommonCrawl.
To our best understanding, our developed dataset is the first of its kind that captures relevant data and signals in an end-to-end ODQA.
In addition, our work also introduces a novel design for optimizing open-domain answer sentence selection tasks operating at Web scale.
Our experimental results show a significant improvement of 6.51\% in accuracy, from 48.86\% to 55.37\% --- while using only $\sim$20\% computation.
We believe our work, the dataset, the models, and the proposed ODQA system for AS2, can benefit several applications that require question-based accurate retrieval and reranking.
We will make our work available to the public community to encourage research and development in this topic.

\bibliography{irqa}
\bibliographystyle{acl_natbib}

\end{document}